# PSO-Optimized Hopfield Neural Network-Based Multipath Routing for Mobile Ad-hoc Networks

**Mansour Sheikhan**

*EE Department, Islamic Azad University, South Tehran Branch, Mahallati Highway, Dah-Haghi Blvd., Tehran, Iran*

*E-mail: msheikhn@azad.ac.ir*

**Ehsan Hemmati**

*EE Department, Islamic Azad University, South Tehran Branch, Mahallati Highway, Dah-Haghi Blvd., Tehran, Iran*

*E-mail: ehemmati@ieee.org*

**Abstract**

Mobile ad-hoc network (MANET) is a dynamic collection of mobile computers without the need for any existing infrastructure. Nodes in a MANET act as hosts and routers. Designing of robust routing algorithms for MANETs is a challenging task. Disjoint multipath routing protocols address this problem and increase the reliability, security and lifetime of network. However, selecting an optimal multipath is an NP-complete problem. In this paper, Hopfield neural network (HNN) which its parameters are optimized by particle swarm optimization (PSO) algorithm is proposed as multipath routing algorithm. Link expiration time (LET) between each two nodes is used as the link reliability estimation metric. This approach can find either node-disjoint or link-disjoint paths in single phase route discovery. Simulation results confirm that PSO-HNN routing algorithm has better performance as compared to backup path set selection algorithm (BPSA) in terms of the path set reliability and number of paths in the set.

*Keywords:* Mobile ad-hoc networks; Reliability; Multipath routing; Neural networks; Particle swarm optimization (PSO)

## 1. Introduction

Mobile ad-hoc networks (MANETs) are defined as the category of wireless networks that utilize multi-hop radio relaying and are capable of operating without the support of any fixed infrastructure. MANETs are useful when no wired link is available such as in disaster recovery or more generally when a fast deployment is necessary. Also, required expensive investments in base stations result to deployment of wireless networks in ad-hoc mode.[1] The tasks such as relaying packets, discovering routes, monitoring the network and securing communication are performed by mobile nodes in the network. Nodes typically communicate in multi-hopping fashion and intermediate nodes act as routers by forwarding data.[2] Unlike the wired networks, route failure is a normal behavior in MANETs. Route failure occurs frequently due to mobility and limited battery power of nodes as well as characteristics of the wireless communication medium. Route recovery process should be done when the route failure occurs in the network. This requires sending extra control packets which consumes network resources like bandwidth and battery power. It also leads to excessive delay that affects the quality of service (QoS) for delay sensitive applications.[3] Routing protocols should adapt to these topology changes and continue to maintain connection between the source and destination nodes in the presence of path breaks caused by link and/or node failures.

In order to increase the routing resistance against link or/and node failures, one solution is to use not just a single path, but a set of redundant paths.[4-9] In this way, there is a fundamental and quite difficult question: which of the potential exponentially many paths within the network should the routing layer use to achieve the highest reliability?

The path with low probability of failure is the reliable one. The correlation of failures between the paths in the set should be as low as possible. Common links and nodes between paths are common failure points in the set. In order to provide high reliable path set, we focus on finding disjoint paths; i.e., the paths with no link- or node- overlap. The problem of finding disjoint paths is non-trivial. Two general principles for selecting the reliable paths can be stated. First, a long path is less reliable than a short one. Second, a larger number of disjoint paths increases the overall reliability. Thus, one should be looking for a large set of short and disjoint paths.

Most of the past works on multipath routing protocols have been based on the single version of an existing routing protocol. They have been mostly focused on load-balancing, delay, energy efficiency and quick failure recovery, but not considered how to effectively select the multiple paths and quality of selected paths as like as the disjointedness of the selected paths. The number of paths found by some algorithms has been restricted to a specific number and they can not select the appropriate number of paths. Also, the proposed algorithms have been limited to find link-disjoint or node-disjoint path set and they are not capable to find both link- and node-disjoint path sets.

Split multipath routing (SMR) algorithm, proposed by Lee and Gerla[8], selects maximally disjoint paths. In SMR, the multipath routes are discovered by a modified route request procedure. In this scheme, the intermediate nodes are allowed to rebroadcast duplicate route request messages if they receive them from a link with better QoS. However in this protocol, the reliability of links has not been used and the paths are not entirely disjoint. It is also limited to route replies provided by the routing protocol. Pearlman et al.[9] have proposed a method which selects the two routes with the least number of hops. This protocol does not provide a metric or model to justify a particular route selection scheme. Selecting paths based on a small number of hops does not imply that paths will undergo less frequent breakages, while the appropriate number of paths may be far from two. Dana et al.[6] have proposed a backup and disjoint path set selection algorithm for MANETs. This algorithm produces a set of backup paths with high reliability. In order to acquire the link reliability estimates, link expiration time (LET) between each two nodes has been used.

The problem of finding the most reliable multipath has already been shown to be computationally hard.[10] It is noted that the motivation for using soft-computing methods is the need to cope with the complexity of existing computational models of real-world systems.[11-15] The recent resurgence of interest in neural networks has its roots in the recognition that human brain performs the computations in a different manner as compared to conventional digital computers. A neural network has a parallel and distributed information processing structure which consists of many processing elements interconnected via weighted connections. One of the important applications of neural network is to solve optimization problems. In these cases, we want to find the best way to do something, subject to certain constraints. The best solution is generally defined by a specific criterion. Hopfield neural network (HNN) is a model that is commonly used to solve optimization and NP-complete problems.[16,17] One of the most important features of this model is that Hopfield network can be easily implemented in hardware, therefore neural computations are performed in parallel and the solution is found more quickly. The use of neural networks to find the shortest path between a given source-destination pair was first introduced by Rauch and Winarske.[18] An adaptive framework to solve the optimal routing problem based on Hopfield neural network has been introduced by Ali and Kamoun.[19]

The computation of the neural network is heavily dependent on the parameters. The parameters should be chosen in such a way that the neural network approaches towards a valid solution.[20] Consequently, tuning the HNN parameters should be done in order to achieve the best solution over the minimum iterations. The lack of clear guidelines in selecting appropriate values of the parameters of energy function is an important issue in the efficiency of HNNs in solving combinatorial optimization problems. It is obvious that a trial and error approach does not ensure the convergence to optimal solutions.[19]

In the recent years, several intelligent optimization algorithms have been used in different applications such as: (a) genetic algorithm (GA) in scheduling problem[21], total cost and allocation problem[22], obtaining the optimal rule set and the membership function for fuzzy-based systems[23], and facility location problem[24], (b) ant colony optimization (ACO) in chaotic synchronization[25] and grouping machines and parts into cells[26], (c) artificial immune method in several nonlinear systems[27], (d) particle swarm optimization (PSO) in single-objective and multi-objective problems[28, 29], bandwidth prediction[30], parameter identification of chaotic systems[31], QoS-aware web service selection in service oriented communication problem[32], and solving multimodal problems[33], (e) harmony search (HS) algorithm for synchronization of discrete-time chaotic systems[34].

Among the mentioned approaches, PSO which has been proposed by Kennedy and Eberhart[35] is inherently continuous and simulates the social behavior of a flock of birds. In PSO, the solution of a specific problem is being represented by multi-dimensional position of a particle and a swarm of particles is working together to search the best position which corresponds to the best problem solution. In each PSO iteration, every particle moves from its original position to a new position based on its velocity. Particle's velocity is influenced by the cognitive and social information of the particles. The cognitive information of a particle is the best position that has been visited by the particle. Based on the traditional speed-displacement search model, Gao et al.[36] have analyzed the PSO mechanism and proposed a generalized PSO model, so that the PSO algorithm can be applied to the fields of discrete and combinatorial optimization.

Bastos-Filho et al.[37] have proposed a swarm intelligence and HNN-based routing algorithm for communication networks. They have used the PSO technique to optimize HNN parameters and the energy function coefficients. The results have shown that the proposed approach achieves better results than existing algorithms that employ the HNN for routing. Hemmati and Sheikhan[38] have proposed a reliable path set selection algorithm based on Hopfield neural network. The performance of their proposed algorithm has been improved by using noisy HNN which introduces more complexity to the HNN implementation[39]. They have shown that the reliability of the multiple disjoint paths found by the proposed algorithm is higher than those found by traditional multipath routing algorithms. But they did not mention how the HNN parameters can be tuned.

In this paper, we introduce a Hopfield neural model to find the most reliable disjoint multipath in a MANET. The proposed scheme contains several parameters and there is no rule to define them exactly. PSO is used to find the best set of parameters used in HNN for multipath calculation. Each node in the network can be equipped with a neural network, and all the network nodes can train and use the neural networks to obtain the optimal or sub-optimal multipath.

In the next section, HNN and PSO are overviewed. The operation and termination conditions of proposed multipath routing algorithm are described in Section 3. The implementation details and the use of PSO to tune the proposed HNN model parameters are presented in Section 4. Section 5 demonstrates the efficiency of the proposed technique through a simulation study. Then, Section 6 conducts a performance evaluation of the proposed algorithm and then the computational complexity of proposed algorithm is described in Section 7. Finally, Section 8 presents the conclusions of the study.

## 2. Background

### 2.1. *Hopfield neural network*

The use of neural networks to solve constrained optimization problems was initiated by Hopfield and Tank[17, 40]. The general structure of HNN is shown in Fig. 1. Assume that the network consists of $n$ neurons. The neurons are modeled as amplifiers in conjunction with resistors and capacitors which compromise feedback circuits. A sigmoid monotonic increasing function relates the output $V_i$ of the $i$th neuron to its input $U_i$:

$$V_i = g(U_i) = \frac{1}{1+e^{-\lambda U_i}}, \qquad (1)$$

where $\lambda$ is a constant called the gain factor. Each amplifier $i$ has an input resistor $r_i$ and an input capacitor $C_i$ which partially define the time constant $\tau_i$ of $i$th neuron. To describe synaptic connections, we can use the matrix $\boldsymbol{T} = [T_{ij}]$, also known as the connection matrix, of the network. A resistor of value $R_{ij}$ connects one of the outputs of the amplifier *j* to the input of amplifier *i*. In this model, each neuron receives an external current (known also as a bias) $I_i$.

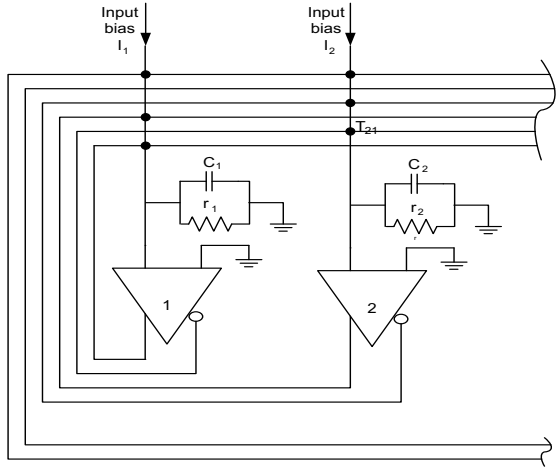

Fig. 1. Hopfield neural network model.

The dynamic of *i*th neuron can be described as follows[16]:

$$\frac{dU_i}{dt} = \sum_{j=1}^{n} T_{ij} V_j - \frac{U_i}{\tau_i} + I_i;$$

$$T_{ij} = \sum_{j=1}^{n} \frac{1}{R_{ij} C_i}, \frac{1}{R_i} = \frac{1}{r_i} + \sum_{j=1}^{n} \frac{1}{R_{ij}}, \tau_i = R_i C_i. \quad (2)$$

For a symmetric connection matrix and for a sufficiently high gain of transfer function, then the dynamics of the neurons follow gradient descent of the quadratic energy function[16]:

$$E = -\frac{1}{2} \sum_{i=1}^{n} \sum_{j=1}^{n} T_{ij} V_i V_j - \sum_{i=1}^{n} I_i V_i. \quad (3)$$

Hopfield has also shown that as long as the state of neural network evolves inside the *N*-dimensional hypercube, defined by $V_i \in \{0,1\}$, if $\lambda_i \to \infty$ the minimum of energy function (3) will attain one of the $2^N$ vertices of this hypercube.

## 2.2. Particle swarm optimization

PSO is a population based stochastic optimization technique which does not use the gradient of the problem being optimized, so it does not require being differentiable for the optimization problem as is necessary in classic optimization algorithms. Therefore it can also be used in optimization problems that are partially irregular, time variable, and noisy.

In PSO algorithm, each bird, referred to as a "particle", represents a possible solution for the problem. Each particle moves through the *D*-dimensional problem space by updating its velocities with the best solution found by itself (cognitive behavior) and the best solution found by any particle in its neighborhood (social behavior). Particles move in a multidimensional search space and each particle has a velocity and a position as follow:

$$v_i(k+1) = v_i(k) + \gamma_{1i}(P_i - x_i(k)) + \gamma_{2i}(G - x_i(k)), \quad (4)$$

$$x_i(k+1) = x_i(k) + v_i(k+1), \quad (5)$$

where *i* is the particle index, *k* is the discrete time index, $v_i$ is the velocity of *i*th particle, $x_i$ is the position of *i*th particle, $P_i$ is the best position found by *i*th particle (personal best), *G* is the best position found by swarm (global best) and $\gamma_{1,2}$ are random numbers in the interval [0,1] applied to *i*th particle. In our simulations, the following equation is used for velocity[41]:

$$v_i(k+1) = \varphi(k) v_i(k) + \alpha_1 [\gamma_{1i}(P_i - x_i(k))] + \alpha_2 [\gamma_{2i}(G - x_i(k))], \quad (6)$$

in which $\phi$ is the inertia function and $\alpha_{1,2}$ are the acceleration constants. The flowchart of the standard PSO algorithm is summarized in Fig. 2.

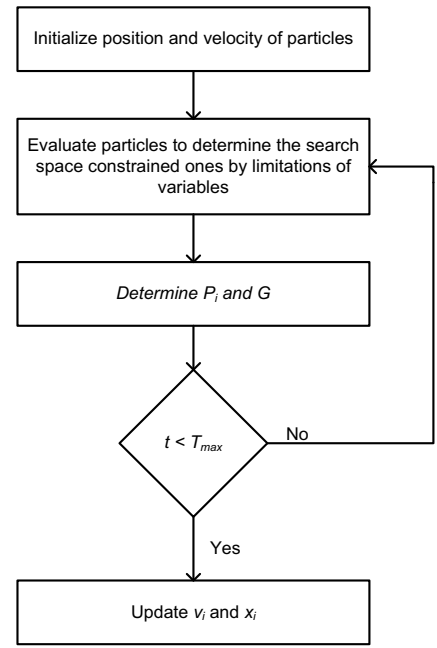

Fig. 2. Standard PSO flowchart.

PSO and the genetic algorithm (GA) are both population-based search algorithms and both of them share information among their population members to enhance their search processes. They also use a combination of deterministic and probabilistic rules. Different experiences have shown that although PSO and GA result on average in the same solution quality, the PSO is more computationally efficient which means it performs less number of function-evaluations as compared to GA. In the other hand, it has been shown that computational effort in PSO and GA is problem dependent; but in solving unconstrained nonlinear problems with continuous design variables, PSO outperforms the GA[42, 43].

As the Hopfield performance is depending on its parameter setting, and it is a continuous problem, PSO is used in this study to determine the optimum values of Hopfield network parameters.

## 3. Proposed Approach

Disjoint multiple paths between source and destination are classified into two types, namely node-disjoint and link-disjoint multiple paths. Node-disjoint paths do not have any nodes in common, except the source and destination. Link-disjoint paths do not have common links, but may share some nodes. Link-disjoint paths are more available than node-disjoint paths. Movement of nodes at the junctions causes the failure of all the paths going through that node. The node-disjoint type has the most disjointedness, as all the nodes/links of two routes are different; i.e., the network resource is exclusive for the respective routes.

Here we propose an algorithm which can compute both node-disjoint and link-disjoint paths. This approach consists of three steps. First a method is introduced to compute the multipath reliability. Then, a route discovery mechanism is defined, and finally the multipath calculation algorithm is proposed in which the most reliable multipath is found by a PSO-optimized HNN.

### 3.1. *Assumptions*

A MANET is denoted by a probabilistic graph $G=(V,L)$, where $V$ is a set of nodes in the network and $L$ is a set of links connecting nodes. Nodes are located on a two-dimensional field and move in the field. Node $i\ (\in V)$ has a distinct identifier $ID_i$. Each node has the wireless transmission range $R > 0$. Node $j$ is called a neighbor of node $i$ if and only if $j$ is within the transmission range $R$ of $i$, and the link $(i,j)$ is included in the link set $L$. The probability of proper operation is also assigned to the links. A link (for example $h$th link) operates with probability $p_h^{link}$ and fails with probability $q_h^{link} = 1 - p_h^{link}$. In this protocol, each node continuously monitors the reliability of its incident links. For each source and destination pair, $Reliability_{S \to D}(S \neq D)$ denotes the probability that there exists at least one path connecting source and destination over $G$ graph.

### 3.2. *Reliability computation method*

Assume that $path_i$, between source and destination, consists of $m$ links. The probability that $path_i$ be operational or the path reliability is obtained by:

$$PathReli_i = \prod_{h=1}^{m} p_h^{link}. \qquad (7)$$

A path fails with probability:

$$PathFail_i = 1 - PathReli_i. \qquad (8)$$

Assume that $P=\{Path_1, Path_2, \ldots, Path_n\}$ denotes a set of disjoint paths that includes $n$ paths. The reliability of the path set is calculated by:

$$\begin{aligned} PSreliability &= 1 - \prod_{i=1}^{n} PathFail_i \\ &= 1 - \prod_{i=1}^{n}(1 - PathReli_i) \\ &\quad 1 - (1 - PathReli_1 - PathReli_2 - \ldots - PathReli_n)\,;\ PathReli_i < 1 \\ &\quad \sum_{i=1}^{n} PathReli_i. \end{aligned} \qquad (9)$$

### 3.3. *Route discovery algorithm*

In this algorithm, each node has a route cache, which preserves the order of nodes and probabilities of all paths, $PathReli_i$, from each source. In order to find paths between the source and destination, the source node broadcasts the route request (RREQ) packet to the nodes which are in its transmission range. This RREQ packet has the following fields:

  *Record*: a record of the sequence of hops taken by this packet;
  *Prob*: the reliability of the followed path;
  *TTL*: the maximum number of hops that a packet can traverse along the network before it is discarded.

When a node receives the RREQ packet, it decrements *TTL* by 1 and performs the following steps:

1. If the node is a destination one, it updates the *Prob,* and adds the *Record* and updates *Prob* to its route cash.

2. If *TTL*=0, the RREQ packet is discarded. Thus, *TTL* limits the number of intermediate nodes in a path.

3. If the ID of this node is already listed in the *Record* of route request, the RREQ packet is discarded to avoid looping.

4. Otherwise, the node appends its node ID to the *Record* in the RREQ packet, and also updates the *Prob* field, and re-broadcasts the request to its neighbors.

When the destination node receives the first RREQ packet from a specific source node, it waits for a while to receive other RREQ packets from longer paths.

Now all the information, which is needed to calculate link-disjoint or node-disjoint paths, is obtained by single route discovery, so there is no need to send extra messages as overhead in the MANET; when both link-disjoint and node-disjoint paths are needed.

For all of the paths in destination route cache, we can assume a disjointedness matrix, $\rho=[\rho_{jk}]$, with the size of $(n \times n)$ in which $n$ is the total number RREQ packets received from a specific source. In order to find node-disjoint paths, we define $ND\rho_{jk}$ as:

$$ND\rho_{jk \atop j \neq k} = \begin{cases} 0 & \text{if } j^{th} \text{ path and } k^{th} \text{ path in} \\ & \text{route cache are node-disjointed} \\ 1 & \text{otherwise.} \end{cases} \quad (10)$$

We define also $LD\rho_{jk}$ to find link-disjoint paths as follows:

$$LD\rho_{jk \atop j \neq k} = \begin{cases} 0 & \text{if } j^{th} \text{ path and } k^{th} \text{ path in} \\ & \text{route cache are link-disjointed} \\ 1 & \text{otherwise.} \end{cases} \quad (11)$$

All the diagonal elements in $\rho$ are also set to zero.

### 3.4. *Neural-based multipath calculation*

To formulate the problem in terms of Hopfield neural model, a suitable representation scheme should be found so that the most reliable multiple disjoint paths can be decoded from the final stable state of the neural network. Each neuron in this model represents a path of discovered paths in the route discovery phase, listed in the route cache. Thus, the total number of neurons required in HNN is equal to the total number of paths found in the route discovery phase. Based on the fact that a path is selected to be in the set or not, the output of a neuron at location *i* is defined as follows:

$$V_i = \begin{cases} 1 & \text{if } i\text{th path in route cache is in the path set} \\ 0 & \text{otherwise.} \end{cases} \quad (12)$$

The normalized reliability of *i*th path is defined as:

$$C_i = \frac{PathReli_i}{PathReli_{max}}, \quad (13)$$

where $PathReli_{max}$ is the highest path reliability along all of the paths in the route cache.

We have to define an energy function whose minimization process drives the neural network into its lowest energy state. This stable state shall correspond to the most reliable set of multiple paths. The energy function must favor states that correspond to disjoint paths and it must also favor the path set which has the highest reliability. A suitable energy function that satisfies such requirements is proposed that is given by:

$$E = \frac{\mu_1}{2} \sum_{i=1}^{n} \sum_{j=1}^{n} \rho_{ij} V_i V_j - \mu_2 \sum_{i=1}^{n} C_i V_i, \quad (14)$$

where $\mu_1$ and $\mu_2$ are positive constants and $n$ is the number of paths in the route cache. In (14), $\rho_{ij}$ is defined as $ND\rho_{ij}$ if we want to calculate node-disjoint multiple paths and it is defined as $LD\rho_{ij}$ if we want to calculate link-disjoint multiple paths. The minimum value of the $\mu_1$ term is zero and it is occurred when all the selected paths are disjoint. The $\mu_2$ term is corresponding to the reliability of the selected multiple disjoint paths. The larger number of high reliable paths results in the lower energy function, but this term should not cause non-disjoint paths participate in the set. By selecting each path from the route cache, the energy decreases by $(\mu_2 \times C_i)$. If the selected path is not disjoint with other paths of the set, then the total energy is changed by $(\mu_1 - \mu_2 \times C_i)$. As we want to select disjoint paths, so this criterion should have higher energy value and $\mu_1 - \mu_2 \times C_i$ should be positive. As $C_i<1$, so $\mu_1$ and $\mu_2$ should meet the criterion $\mu_1 \geq \mu_2 > 0$. In this way, assume that $P_1 = \{Path_1, Path_2, ..., Path_k\}$ is a set of disjoint paths and $P_2 = P_1 \cup \{Path_l\}$ which $Path_l \notin P_1$ has a common node (or link) with at least one path in $P_1$. Also assume that $E_{P1}$ is the energy of $P_1$ and $E_{P2}$ is the energy of $P_2$. Based on these assumptions, we can write $E_{P2}=E_{P1}+\mu_1-\mu_2 \times C_i$. As $P_2$ is not disjoint, so its energy should be higher than the energy of $P_1$. In other words, $E_{P2}>E_{P1}$; thus $\mu_1-\mu_2 \times C_l > 0$. From (13) it is obvious that $C_l \leq 1$, so in order to keep $E_{P2}>E_{P1}$ always true, $\mu_1$ and $\mu_2$ should meet the criterion $\mu_1 \geq \mu_2 > 0$.

By comparing the corresponding coefficients in (14) and (3), the connection strengths and the biases are derived by:

$$T_{ij} = -\mu_1 \rho_{ij}$$
$$I_i = \mu_2 C_i. \tag{15}$$

As can be seen in (15), this model maps the reliability information into the biases and path disjointedness information of the neural interconnections. So the destination node can set the neural interconnection as it received each RREQ packet and then set the biases after all the RREQ packets has been received. The destination node uses this neural network to find the most reliable path set, and then returns a copy of neural network solution in a route reply packet to the source node.

The conceptual scheme of implementing the proposed algorithm in a MANET node is depicted in Fig. 3. According to the network properties, the neural network parameters can be tuned by PSO once and then these parameters can be used.

## 4. Implementation Details and Parameter Tuning

In order to implement this algorithm, there should be a method to predict the link reliability and also a network model that dictates how the nodes move throughout the network and structure of the network itself. We also need to define the Hopfield implementation method.

### 4.1. Link reliability prediction

We consider a free space propagation model[44]. In this model, the wireless signal strength depends only on the distance to the transmitter. Hence, the link duration of $L_{ij}$ can be predicted from the motion information of the two nodes. Assume that node $i$ and node $j$ are within the same transmission range, $r$, of each other. Let $(x_i,y_i)$ be the coordinate of node $i$ and $(x_j,y_j)$ be that of node $j$. Also let $v_i$ and $v_j$ be the speeds, and $\theta_i$ and $\theta_j$ ($0 \leq \theta_i, \theta_j < 2\pi$) be the moving directions of node $i$ and $j$, respectively. Then, the amount of time that the two mobile hosts will stay connected, LET, is predicted as follows[45]:

$$LET(i,j) = \frac{-(ab+cd)+\sqrt{(a^2+c^2)r^2-(ad-bc)^2}}{a^2+c^2};$$
$$b = x_i - x_j,$$
$$d = y_i - y_j, \tag{16}$$
$$a = v_i \cos\theta_i - v_j \cos\theta_j,$$
$$c = v_i \sin\theta_i - v_j \sin\theta_j.$$

The probability of proper operation of $h$th link (the link between node $i$ and node $j$) is calculated by:

$$p_h^{link} = \frac{LET_{i,j}}{LET_{max}}, \tag{17}$$

where $LET_{max}$ is the maximum link expiration time in the network.

### 4.2. Ad-hoc network model

All of the nodes start the experiment at a random location within a rectangular working area of 1000×500 m$^2$ and moved as defined by the random waypoint model[46]. For this, each node selects a random destination within the working area and moves linearly to that location at a predefined speed. After reaching its destination, it pauses for a specified time period (pause time) and then the node selects a new random location and continues the process again.

In the present study, each node pauses at the current position for 5 s and speed of individual nodes ranges from 0 to 20 m/s. We have run simulations for a network with 30 mobile hosts, operating at transmission ranges ($R$) varying from 100 to 300 m. The *TTL* is set to 3 as an initial value. The proposed model consists of $n$ neurons where $n$ is the total number of paths in the destination route cache found in the route discovery phase. The average paths found in the route discovery phase in different MANETs is calculated where *TTL*=3. The result is shown in Fig. 4.

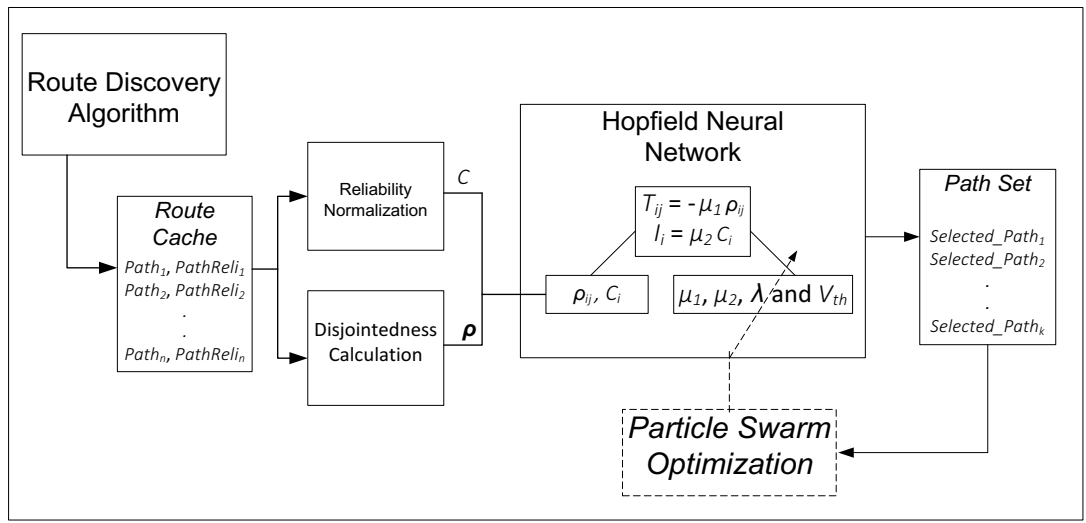

Fig. 3. Conceptual scheme of proposed algorithm.

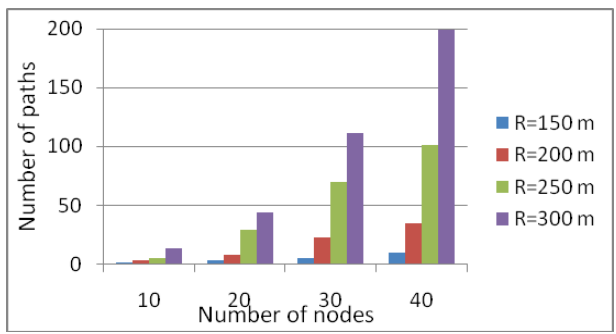

Fig. 4. Number of paths averaged over different MANETs.

### 4.3. Hopfield network initialization

The evolution of the neural network state is simulated by the solution of a system of *n* differential equations where the variables are the neuron outputs ($V_i$). The solution consists of observing the outputs $V_i$ for a specific duration $\delta t$. Without loss of generality, it has been considered that $\tau=1$. To avoid bias in favor of any particular path set, it is assumed that all the inputs $U_i$ are equal to 0. However, to help rapid convergence of the network, small perturbations are applied to the initial inputs of network. Initial random noise helps to break symmetry which may caused by paths with same reliability or by the possibility of having two or more high reliable path set, while preventing it from adopting an undesirable state[20, 47]. At the start and based on our simulation results, $U_i$'s are chosen randomly such that $-0.0005<U_i<0.0005$. The calculations are stopped when the network reaches a stable state; e.g., when the difference between the outputs is less than $10^{-6}$ from one update to another. When the network is in a stable state, the final values of $V_i$ are rounded off. For this reason, the threshold voltage, $V_{th}$, should be defined and $V_i$ should be rounded off in such a way that it is set to 0 if $V_i<V_{th}$, and to 1 otherwise.

### 4.4. Selecting network parameters by PSO

The PSO algorithm is used to find the values of $\mu_1$, $\mu_2$, $\lambda$, $\delta t$ and $V_{th}$ of the HNN. Each dimension of the PSO particle is used to present a different HNN parameter, thus each particle has five dimensions. To evaluate the fitness of each particle, we compute the percentage error obtained by 500 HNN simulations. An error is assumed to have occurred if the HNN method finds multiple paths which are not disjoint or the reliability of the set is less than reliability of the set found by parameter setting reported in Ref. 38 (Table 1). The initial values of PSO parameters are shown in Table 2. Table 3 depicts the maximum value (*Xmax*) and the minimum value (*Xmin*) of the parameters. The result of applying PSO algorithm to obtain the optimum values for HNN parameters is shown in Table 4.

Table 1. Values of HNN parameters reported in Ref. 38.

| Parameter | $\mu_1$ | $\mu_2$ | $\lambda$ | $\delta t$ | $V_{th}$ |
|---|---|---|---|---|---|
| Value | 1 | 1 | 50 | $10^{-5}$ | 0.1 |

Table 2. Initial values of PSO parameters.

| Parameter | Max. No. of iterations | Population size | Max. particle velocity | Initial inertia weight | Final inertia weight | Min. global error gradient |
|---|---|---|---|---|---|---|
| Value | 300 | 20 | 4 | 0.9 | 0.2 | $10^{-5}$ |

Table 3. Maximum and minimum values of the parameters.

| Parameter | $\mu_1$ | | $\mu_2$ | | $\lambda$ | | $\delta t$ | | $V_{th}$ | |
|---|---|---|---|---|---|---|---|---|---|---|
| | Xmax | Xmin | Xmax | Xmin | Xmax | Xmin | Xmax | Xmin | Xmax | Xmin |
| Value | 50 | 0 | 50 | 0 | 100 | 0 | 1 | 0 | 1 | 0 |

Table 4. Optimum parameter values obtained by PSO.

| Parameter | $\mu_1$ | $\mu_2$ | $\lambda$ | $\delta t$ | $V_{th}$ |
|---|---|---|---|---|---|
| Value | 32 | 27 | 0.45 | $10^{-3}$ | 0.23 |

## 5. Simulation Results

For the purpose of evaluating the efficiency of proposed routing method, it has been applied to different networks with various parameters. Then, the total path reliability, the number of paths in path set, and the lifetime are calculated for each simulation. Several MANETs with different characteristics are considered in this study in order to tune the HNN parameters and the simulation results are shown that HNN with this parameter setting have good performance when applying to a variety of MANETs. Therefore, the tuned parameters obtained in this study can be used in other MANETs also, or it can be tuned based on that MANET once and then using the same parameter for all of the nodes in a network.

Lifetime is considered as the time between the construction of path set and the breakage of all paths in the path set. Based on the network parameters, two different scenarios have been considered. In each scenario, one aspect of MANET characteristics is considered. For example in the first scenario, the connectivity is considered and the network density is considered in the second scenario. Then the simulation results for these scenarios, using the proposed routing algorithm, are compared with previous works, non-optimized HNN and noisy HNN path set selection algorithms reported in Refs. 38 and 39. All the simulation programs have been written and compiled in MATLAB 7.10 and run on PC with Intel Pentium E5300 CPU and 2 GB RAM.

In the first scenario, the number and speed of nodes are considered fixed and the transmission range is variable. Since in this scenario, the transmission range is variable the focus of this scenario is to study the effect of this parameter on the reliability and number of paths. As depicted in Fig. 5, the reliability of disjoint paths selected by PSO-optimized HNN is higher than two other algorithms. By comparing the reliability between link-disjoint paths (Fig. 5a) and node-disjoint paths (Fig. 5b), we find that link-disjoint paths are more reliable than node-disjoint paths in the same transmission range, because there are more choices to select the link-disjoint paths rather than the node-disjoint paths. Consequently, if the transmission range is increased, then the number of selected paths is also increased for both sets of link-disjoint and node-disjoint paths (Fig. 6). When the radio transmission range of nodes increases, there will be more paths between source and destination nodes which routing algorithm can select among them.

The time between the construction of path set and the breakage of all paths in the path set is called lifetime or time to failure. As shown in Fig. 7, the lifetime is increased as the transmission range increases.

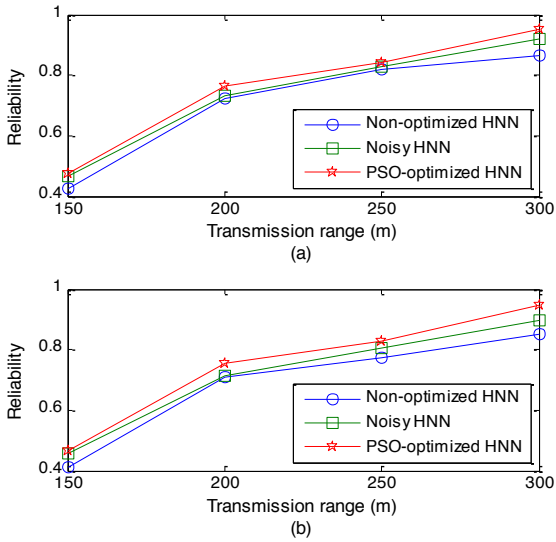

Fig. 5. Reliability of different HNN-based path selection algorithms; a) Link-disjoint, b) Node-disjoint.

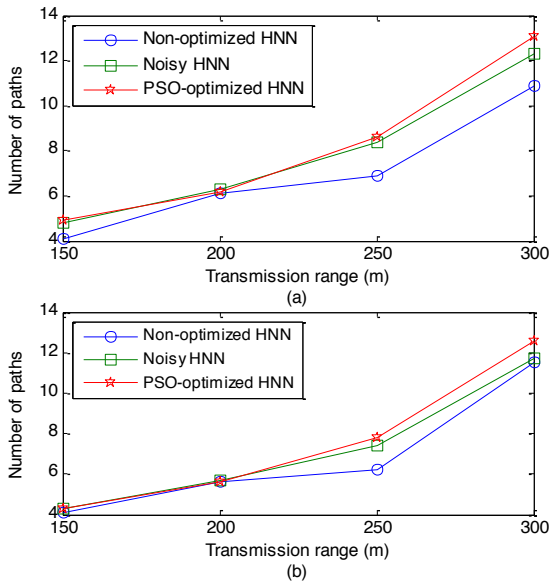

Fig. 6. Number of paths selected by different HNN-based algorithms; a) Link-disjoint, b) Node-disjoint.

Link-disjoint path sets have longer lifetime than node-disjoint ones. In this case, the PSO-optimized HNN algorithm also shows better performance than others which means the disjoint paths selected by this algorithm are more reliable and the connection can continue longer by using the paths selected by this algorithm.

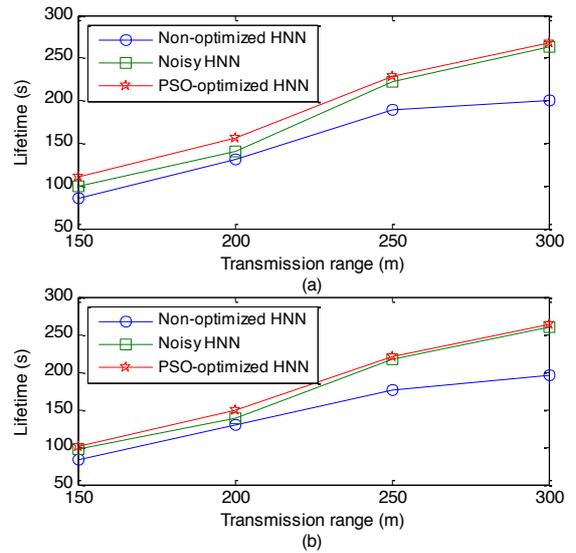

Fig. 7. Lifetime of selected paths by different HNN-based algorithms; a) Link-disjoint, b) Node-disjoint.

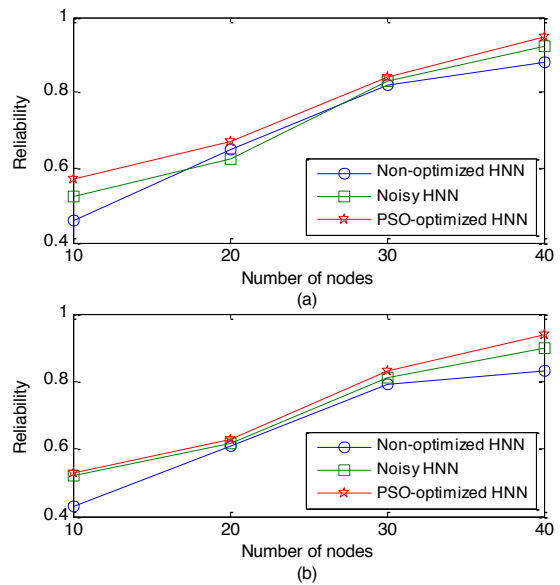

Fig. 8. Reliability for different number of nodes using HNN-based algorithms; a) Link-disjoint paths, b) Node-disjoint paths.

The second scenario considers a fixed transmission range while the number of nodes is variable. In this scenario, the transmission range of all nodes is set to 250 m. The reliability of the set for both the node-disjoint and the link-disjoint paths is shown in Fig. 8. As can be seen, in a high density network, there are more routes in the path set and the path set reliability is higher than those of a low density network.

Although the PSO-optimized HNN has similar or only slightly better results with those of a noisy HNN, but it is noticeable that the implementation of the HNN is simpler than a noisy HNN. It can be found from the simulation results that by fine tuning of the HNN, better results can be achieved with a simpler implementation and there is no need of an additional hardware.

## 6. Performance Evaluation

A performance comparison between the proposed algorithm and the shortest path (SP) algorithm is done in this study and the results are listed in Table 5. As can be seen, the reliability of path set in the proposed algorithm outperforms the SP algorithm. The best improvement in reliability is achieved when the transmission range is 250 m. In this transmission range, the link-disjoint path set reliability is 4.5 times more than the corresponding shortest path reliability. Also, the best improvement in lifetime is achieved when the transmission range is 250 m. In this transmission range, the link-disjoint path set lifetime is 3.2 times more than the corresponding shortest path lifetime.

Hemmati and Sheikhan[38] have proposed a method for path set selection using Hopfield neural network. Hopfield parameter settings in Ref. 38 have been based on the values reported in Table 1. The average number of iterations for both PSO-optimized and non-optimized settings are reported in Table 6. The PSO-optimized HNN which is reported in this study takes less iterations and thus less time to reach the steady state and get the solution. The reliability of path sets found by PSO-optimized is also higher than those found by non-optimized HNN.

Dana et al.[6] have proposed the backup path set selection algorithm (BPSA), based on a heuristic and picks a set of highly reliable paths. As the BPSA can find just the link-disjoint path set, we compare the link-disjoint path set, selected by PSO-optimized HNN algorithm with those selected by BPSA algorithm. Table 7 shows the comparison between the proposed algorithm and BPSA. The values of path set reliability and number of paths are averaged over several simulations with different MANETs. The proposed algorithm has better performance in both the reliability and the number of paths. It shows up to 58.3% improvement in the path set reliability and up to 22.4% improvement in the number of paths in the set.

## 7. Computational Complexity of Algorithm

To determine the computational complexity, three components should be considered: route discovery calculations, calculating the elements of $\rho$ matrix, and normalized reliability calculations.

To determine this complexity, assume that *M=TTL*. The maximum number of nodes in each path of the proposed algorithm is $M+1$, in which source and destination are included. So, there is $M-1$ intermediate nodes that the algorithm finds them in the route discovery phase. At the worst case, where all of the MANET nodes are in the transmission range of each other, there are $O(|V|^{(M-1)})$ such routes.

Table 5. Reliability and lifetime comparison between proposed algorithm and SP algorithm.

| Transmission range (m) | Reliability | | | Lifetime (s) | | |
|---|---|---|---|---|---|---|
| | Node-disjoint | Link-disjoint | SP | Node-disjoint | Link-disjoint | SP |
| 150 | 0.466 | 0.475 | 0.107 | 101.3 | 111.5 | 48.3 |
| 200 | 0.754 | 0.762 | 0.171 | 149.2 | 155.9 | 62.4 |
| 250 | 0.839 | 0.865 | 0.191 | 221.7 | 228.8 | 71.2 |
| 300 | 0.949 | 0.951 | 0.352 | 264.9 | 267.3 | 87.5 |

Table 6. Performance comparison between PSO-optimized and non-optimized HNN algorithms.

| Transmission range (m) | Number of iterations | | | Reliability | | |
|---|---|---|---|---|---|---|
| | Non-optimized | PSO-optimized | Percent of iterations in optimized to non-optimized model | Non-optimized | PSO-optimized | Percent of increment in optimized as compared to non-optimized model |
| 150 | 70476 | 8587 | 12.2 | 0.415 | 0.470 | 13.3 |
| 200 | 87944 | 7896 | 9.0 | 0.723 | 0.758 | 4.8 |
| 250 | 83969 | 9601 | 11.4 | 0.806 | 0.837 | 3.8 |
| 300 | 95040 | 11312 | 11.9 | 0.855 | 0.950 | 11.1 |

Table 7. Performance comparison between proposed algorithm and BPSA algorithm.

| Transmission range (m) | Reliability | | Number of paths | |
|---|---|---|---|---|
| | Proposed algorithm | BPSA algorithm | Proposed algorithm | BPSA algorithm |
| 150 | 0.47 | 0.38 | 4.9 | 4.8 |
| 200 | 0.76 | 0.48 | 6.2 | 6.1 |
| 250 | 0.84 | 0.67 | 8.6 | 7.4 |
| 300 | 0.95 | 0.71 | 13.1 | 10.7 |

To determine the route discovery computational complexity, the complexity in path reliability calculations should be considered. In a naive implementation, the path reliability is calculated independently for each route. The maximum length of each path between two nodes in the MANET is $M$, so the route discovery algorithm has to make $O(M |V|^{(M-1)})$ calculations. To calculate the elements of $\rho$ matrix, $(M-1)^2$ comparisons for node-disjoint and $M^2$ comparisons for link-disjoint path sets should be performed between each two paths of total $O(|V|^{(M-1)})$ paths to determine $\rho$ matrix. So, $\rho$ matrix calculation for the mentioned path sets needs $O(|V|^{2(M-1)})$ operations.

To calculate the normalized path reliability ($C_i$), it is noted that according to (13) the number of operations in this part is equal to the total number of paths. The computational complexity of this part is $O(|V|^{(M-1)})$.

Since the computation time in neural networks is expected to be very short, then the complexity of the proposed approach for the path set selection is best assessed in terms of the programming complexity, which is defined as the number of arithmetic operations required to determine again the proper synaptic connections and the biases each time a new data is fed to the neural net. According to (15), we can conclude that by determining the elements of $\rho$ matrix, the synaptic connections of neural network ($T_{ij}$) are also calculated. The biases are also specified when the normalized path reliabilities ($C_i$) are determined. In this study, we set $M$ to 3 so the total computational complexity in the worst case is $O(|V|^4)$.

## 8. Conclusion

In this paper, we have proposed a reliable multipath routing algorithm using Hopfield neural network optimized by PSO for MANETs. A reliable path is constructed by the links that keep connection between two nodes for a long time. Each node predicts disconnection of all the incident links. A flooding mechanism has been used for the route discovery. Finding multiple paths in a single route discovery in this algorithm reduces the routing overhead. The proposed algorithm is able to compute both link-disjoint and node-disjoint multiple paths. Multipath routing in MANET consists of determining the most reliable disjoint multiple paths between each node pair within the network. The disjoint multiple path selection algorithm is proposed using Hopfield neural network. In order to improve the network performance, PSO algorithm is used to optimize the HNN parameters.

The simulation results show that PSO is a reliable approach to optimize the Hopfield network for multipath routing, since this method results in fast convergence and produces more accurate results as compared to non-optimized HNN, noisy HNN, shortest path (SP) algorithm and recent researches in this field. The simulation results have shown that for different network conditions, the proposed model is efficient in selecting multiple disjoint paths. Simulations also have shown that the link-disjoint path set is more reliable than the node-disjoint one in different conditions.

Simulation results show that the reliability and lifetime are increased up to 4.5 and 3.2 times as compared to shortest path routing algorithm, respectively. The PSO-optimized HNN routing algorithm has better performance as the reliability of multiple paths is increased by 8.3%, while the number of algorithm iterations is reduced to 11.1% as compared to the non-optimized HNN multipath routing when averaged over different transmission ranges. Also, the proposed algorithm has better performance in terms of reliability and number of paths when compared with the backup path set selection algorithm (BPSA). In this way, it shows up to 58% improvement in the path set reliability and up to 22% improvement in the number of paths in the set.